\def\year{2018}\relax
\documentclass[letterpaper]{article} 
\usepackage{aaai18}  
\usepackage{times}  
\usepackage{helvet}  
\usepackage{courier}  
\usepackage{url}  
\usepackage{graphicx}  
\usepackage{comment}

\usepackage{subfig}
\usepackage{booktabs} 
\usepackage{balance}  
\usepackage{subfig}
\usepackage[table]{xcolor}
\definecolor{high}{HTML}{FFCE8E}

\begin{document}
%
\title{Hate Lingo: A Target-based Linguistic Analysis of Hate Speech in Social Media}
\author{Mai ElSherief, Vivek Kulkarni, Dana Nguyen, William Yang Wang, Elizabeth Belding \\
University of California, Santa Barbara\\
{\{mayelsherif, vvkulkarni, dananguyen, william, ebelding\}}@ucsb.edu
}
\maketitle
\begin{abstract}
While social media empowers freedom of expression and individual voices, it also enables anti-social behavior, online harassment, cyberbullying, and hate speech. In this paper, we deepen our understanding of online hate speech by focusing on a largely neglected but crucial aspect of hate speech -- its \emph{target}: either \emph{directed} towards a specific person or entity, or \emph{generalized} towards a group of people sharing a common protected characteristic. We perform the first linguistic and psycholinguistic analysis of these two forms of hate speech and reveal the presence of interesting markers that distinguish these types of hate speech. Our analysis reveals that Directed hate speech, in addition to being more personal and directed, is more informal, angrier, and often explicitly attacks the target (via name calling) with fewer analytic words and more words suggesting authority and influence. Generalized hate speech, on the other hand, is dominated by religious hate, is characterized by the use of lethal words such as murder, exterminate, and kill; and quantity words such as million and many. Altogether, our work provides a data-driven analysis of the nuances of online-hate speech that enables not only a deepened understanding of hate speech and its social implications, but also its  detection.  


\end{abstract}

\section{Introduction}
Social media is an integral part of daily lives, easily facilitating communication and exchange of  points of view. On one hand, it enables people to share information, provides a framework for support during a crisis \cite{olteanu2015expect}, aids law enforcement agencies \cite{crump2011police} and more generally facilitates insight into society at large. On the other hand, it has also opened the doors to the proliferation of anti-social behavior including online harassment, stalking, trolling, cyber-bullying, and hate speech. 
In a Pew Research Center study\footnote{http://www.pewinternet.org/2014/10/22/online-harassment/}, 60\% of Internet users said they had witnessed offensive name calling, 25\% had seen someone physically threatened, and 24\% witnessed someone being harassed for a sustained period of time. 
Consequently, \emph{hate speech} -- speech that denigrates a person because of their innate and protected characteristics -- has become a critical focus of research. 

\begin{figure}[ht!]
{\centering \includegraphics[width = 0.47\textwidth, height = 4cm]{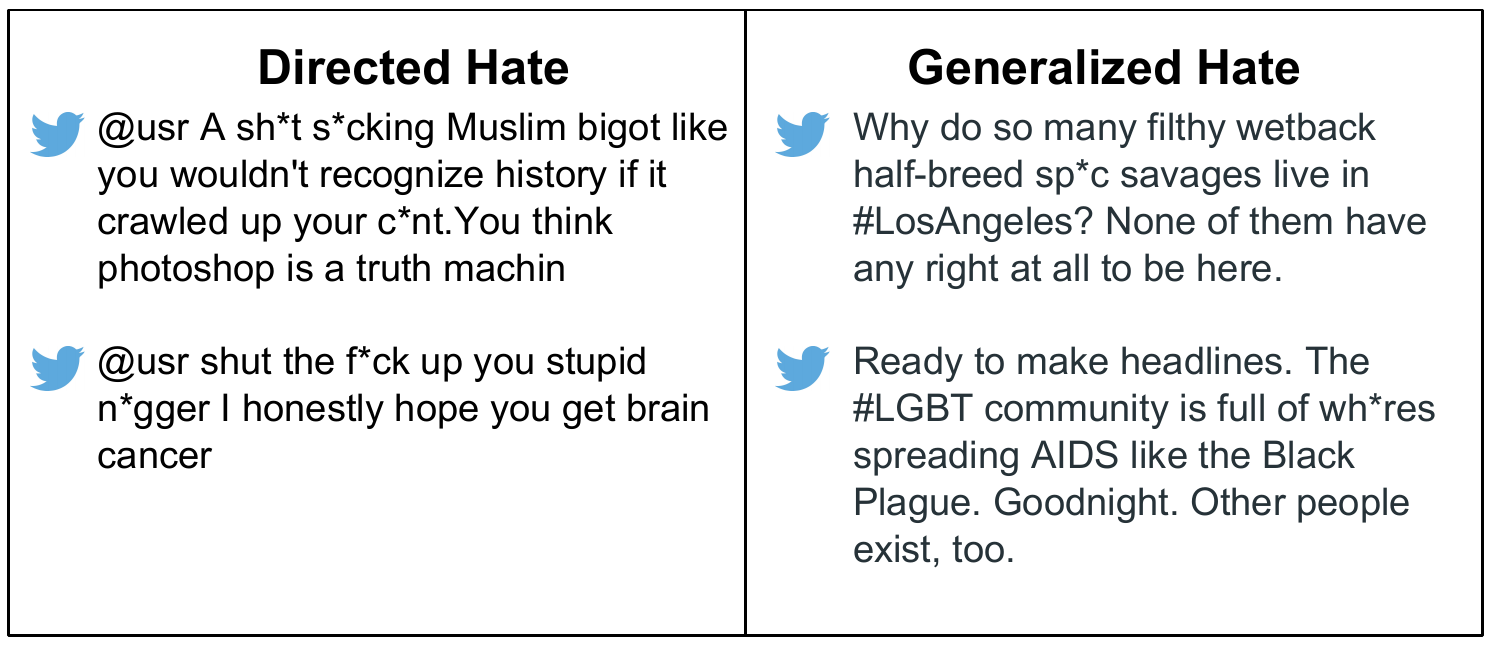}}
\caption{Examples of two different types of hate speech. Directed hate speech is explicitly directed at an individual entity while Generalized hate speech targets a particular community or group.  Note that throughout the paper, explicit text has been modified to include a star (*).}
\label{fig:crown_jewel}
\end{figure}

However, prior work ignores a crucial aspect of hate speech -- the \emph{target of hate speech} -- and only seeks to distinguish \emph{hate} and \emph{non-hate speech}. Such a binary distinction fails to capture the nuances of hate speech -- nuances that can influence free speech policy. First, hate speech can be directed at a specific individual (\textbf{Directed}) or it can be directed at a group or class of people (\textbf{Generalized}). Figure~\ref{fig:crown_jewel} provides an example of each hate speech type. Second, the target of hate speech can have legal implications with regards to right to free speech (the First Amendment).\footnote{We refer the reader  to ~\cite{wolfson1997hate} for a detailed discussion of one such case and its implications.} 

In this work, we bridge the gaps identified above by analyzing Directed and Generalized hate speech to provide a thorough characterization. Our analysis reveals several differences between \textbf{Directed} and \textbf{Generalized} hate speech. First, we observe that Directed hate speech is very personal, in contrast to Generalized hate speech, where religious and ethnic terms dominate. Further, we observe that generalized hate speech is dominated by hate towards religions as opposed to other categories, such as Nationality, Gender or Sexual Orientation. We also observe key differences in the linguistic patterns, such as the semantic frames, evoked in these two types. More specifically, we note that \textbf{Directed} hate speech invokes words that suggest \emph{intentional action}, \emph{make statements} and explicitly uses words to hinder the action of the target (e.g. calling the target a \texttt{retard}). In contrast, \textbf{Generalized} hate speech is dominated by \emph{quantity words} such as \texttt{million, all, many}, \emph{religious words} such as \texttt{Muslims, Jews, Christians} and \emph{lethal words} such as \texttt{murder, beheaded, killed, exterminate}.  
Finally, our psycholinguistic analysis reveals language markers suggesting differences between the two categories. One key implication of our analysis suggests that \textbf{Directed} hate speech is more informal, angrier and indicates higher clout than \textbf{Generalized} hate speech. Altogether, our analysis sheds light on the types of digital hate speech, and their distinguishing characteristics, and paves the way for future research seeking to improve our understanding of hate speech, its detection and its larger implication to society. This paper presents the following contributions:
\begin{itemize}
\item We present the first extensive study that explores different forms of hate speech based on the target of hate. 
\item We study the lexical and semantic properties characterizing both \textbf{Directed} and \textbf{Generalized} hate speech and reveal key linguistic and psycholinguistic patterns that distinguish these two types of hate speech. 
\item We curate and contribute a dataset of 28,318 Directed hate speech tweets and 331 Generalized hate speech tweets to the existing public hate speech corpus.\footnote{The datasets are available here: \url{https://github.com/mayelsherif/hate_speech_icwsm18}}

\end{itemize}

%

\section{Related Work}

\textbf{Anti-social behavior detection.} In $1997$, the use of machine learning was proposed to detect classes of abusive messages~\cite{spertus1997smokey}. 
Cyberbullying has been studied on numerous social media platforms, \textit{e.g.,} Twitter~\cite{burnap2015cyber,silva2016analyzing} and YouTube~\cite{dinakar2012common}. 
Other work has focused on detecting personal insults 
and offensive language~\cite{huang2013using,burnap2015cyber}. 


A proposed solution for mitigating hate speech is 
to design automated detection tools  with social content moderation. 
A recent survey outlined eight categories of features used in hate speech detection~\cite{schmidt2017survey} including simple surface, word generalization, sentiment analysis, lexical resources and linguistic features, knowledge-based features, meta-information, and multi-modal information.

\textbf{Hate speech detection.}
Hate speech detection has been supplemented by a variety of features including lexical properties such as n-gram features~\cite{nobata2016abusive}, character n-gram features~\cite{mehdad2016characters}, average word embeddings, and paragraph embeddings~\cite{nobata2016abusive,djuric2015hate}. Other work has leveraged sentiment markers, specifically negative polarity and sentiment strength in preprocessing~\cite{dinakar2012common,sood2012automatic,gitari2015lexicon} and as features for hate speech classification~\cite{van2015detection,burnap2015tension}. In contrast, our work reveals novel linguistic, psychological, and affective features inferred using an open vocabulary approach to characterize Directed and Generalized hate speech. 


\textbf{Hate speech targets.}
Silva~\textit{et al.} study the targets of online speech by searching for sentence structures similar to ``I $<$intensity$>$ hate $<$targeted group$>$''. They find that the top targeted groups are primarily bullied for their ethnicity, behavior, physical characteristics, sexual orientation, class, or gender. Similar to~\cite{silva2016analyzing}, we differentiate between hate speech based on the innate characteristic of targets, \textit{e.g.,} class and ethnicity. However, when we collect our datasets, we use a set of diverse techniques and do not limit our curation to a specific sentence structure.

\section{Data, Definitions and Measures}
We adopt the definition of hate speech along the same lines of prior literature~\cite{hine2017kek,davidson2017automated} and inspired by social networking community standards and hateful conduct policy~\cite{fb2017contrharmhatespeech,twitter2017hatefulconductpolicy} as ``\textit{direct and serious attacks on any protected category of people based on their race, ethnicity, national origin, religion, sex, gender, sexual orientation, disability or disease}''. 
~\citeauthor{waseem2017understanding} outline a typology of abuse language and differentiate between Directed and Generalized language. We adopt the same typology and define the following in the context of hate speech: 
\begin{itemize}
\item \textbf{Directed hate}: hate language towards a specific individual or entity. An example is:
\textit{``@usr\footnote{Note that we anonymize all user mentions by replacing them with \textit{@usr}.} your a f*cking queer f*gg*t b*tch''}.
\item \textbf{Generalized hate}: hate language towards a general group of individuals who share a common protected characteristic, e.g., ethnicity or sexual orientation. An example is: \textit{``| was born a racist and | will die a racist! | will not rest until every worthless n*gger is rounded up and hung, n*ggers are the scum of the earth!! wPww WHITE America''}.
\end{itemize}

\begin{table*}[!tb]
\footnotesize
\centering 
\resizebox{\textwidth}{!}{%

\begin{tabular}{l||c|c|c|c|c||c|c||c}
\textbf{Category} & \textbf{Key phrase-based} & \textbf{Hashtag-based} & \textbf{Davidson~\textit{et al.}} & \textbf{Waseem~\textit{et al.}} & \textbf{NHSM} & \textbf{Generalized} & \textbf{Directed} & \textbf{Gen-1\%}  \\\hline
Archaic & 169 & 0 & 7 & 0 & 0 & 5 & 171 & - \\
Class & 917 & 0 & 138 & 0 & 0 & 107 & 948 & -\\
Disability & 8,059 & 0 & 63 & 0 & 0 & 35 & 8,087 & -\\
Ethnicity & 2,083 & 220 & 617 & 0 & 16 & 648 & 2,288 & -\\
Gender & 13,272 & 0 & 58 & 0 & 2 & 43 & 13,289 & -\\
Nationality & 81 & 0 & 4 & 0 & 5 & 8 & 83 & - \\
Religion & 48 & 70 & 46 & 1,651 & 9 & 1444 & 380 & -\\
Sexorient & 3,689 & 0 & 394 & 0 & 9 & 253 & 3,840 & -\\\hline
\rowcolor{high}
Total & 28,318 & 290 & 1,327 & 1,651 & 41 & 2,543 & 29,086 & 85,000\\
\hline
\end{tabular}
}
\caption{Categorization of all collected datasets.}
\label{all-dataset}
\end{table*}



\subsection{Data and Methods}
\label{datasets}
Despite the existence of a body of work dedicated to 
detecting hate speech~\cite{schmidt2017survey}, accurate hate speech detection is still extremely challenging~\cite{cnn2016twitter}. A key problem is the lack of a commonly accepted benchmark corpus for the task. Each classifier is tested on a corpus of labeled comments ranging from a hundred to several thousand~\cite{dinakar2012common,van2015detection,djuric2015hate}.
Despite the presence of public crowdsourced slur databases~\cite{trsdb1999theracialslurdb,swear2017swearwordlist}, filters and classifiers based on specific hate terms have proven to be unreliable since (i) malicious users often use misspellings and abbreviations to avoid classifiers~\cite{sood2012profanity}; (ii) many keywords can be used in different contexts, both benign and hateful; and (iii) the interpretation or severity of hate terms can vary based on community tolerance and contextual attributes. 
Another option for collecting a dataset is filtering comments based on hate terms and annotating them. This is  challenging because (i) annotation is time consuming and the percentage of hate tweets is very small relative to the total; and (ii) there is no consensus on the definition of hate speech~\cite{sellars2016defining}. Some work has distinguished between profanity, insults and hate speech~\cite{davidson2017automated}, while other work has considered any insult based on the intrinsic characteristics of the person (e.g. ethnicity,  sexual orientation, gender) to be hate speech related~\cite{warner2012detecting}. To mitigate the aforementioned challenges we adopt several strategies including a comprehensive human evaluation. We describe the construction of our datasets below in detail.  The datasets themselves are summarized in Table~\ref{all-dataset}.

\textbf{(1) Key phrase-based dataset:} We adopt a multi-step classification approach. First, we use Twitter's Streaming API\footnote{Twitter Streaming APIs: https://dev.twitter.com/streaming/overview} to procure a 1\% sample of Twitter's public stream from January 1st, 2016 to July 31st, 2017.  We use Hatebase\footnote{Hatebase: https://www.hatebase.org/}, the world's largest online repository of structured, multilingual, usage-based hate speech as a lexical resource to retrieve English hate terms\footnote{We refer to hate speech terms as keyphrases, keywords, hate terms and hate expressions.}, broken down as: 42 archaic terms, 57 class, 7 disability, 427 ethnicity, 13 gender, 147 nationality-related, 38 religion, and 9 related to sexual orientation. After careful inspection and five iterations of keyword scrutiny by human experts, we removed keyphrases that resulted in tweets with uses distinct from hate speech or phrases that were extremely context sensitive. For example, the word ``pancake'' appears in Hatebase, but clearly can be used in benign contexts. Since our goal was a high quality dataset, we only included keyphrases that were highly likely to indicate hate speech. 

Despite the qualitative inspection of the keyphrases, when we used the resultant keyphrases to filter tweets from the 1\% public stream, non-hate speech tweets remained in our dataset. As an example, speech denouncing hate speech was incorrectly categorized as hate speech. For example, consider the following two tweets: \\
(a): ``\textit{@usr\_1 i'll tear your limbs apart and feed them to the f*cking sharks you n*gger}'' \\
(b): ``\textit{@usr\_2 what influence?? that you can say n*gger and get away with it if you say sorry??}. \\
While both of these tweets contain the word ``n*gger'', the first tweet (a) is pro-hate speech where the hate instigator is attacking \textit{usr\_1}; the second tweet (b) is anti-hate speech in which the tweet author denounces the comments of \textit{usr\_2}. Thus stance detection is vital to consider when classifying hate speech tweets. To mitigate the effects of obscure contexts and stance with respect to hate speech on the filtering process, we used  the Perspective API\footnote{Conversation AI source code: https://conversationai.github.io/} developed by Jigsaw and the Google Counter-Abuse technology team, the model behind which is comprehensively discussed in ~\cite{wulczyn2017ex}.\footnote{We also experimented with classifiers including~\cite{davidson2017automated} but found Perspective API to be empirically better.}
The Perspective API contains different models of classification including: toxicity, attack of commenter, inflammatory, and obscene, among others. When a request is sent to the API with specific model parameters, a probability value [0, 1] is returned for each model type. For our datasets, we focus on two models: \texttt{toxicity} and \texttt{attack\_on\_commenter} models. The \texttt{toxicity} model is a convolutional neural network trained with word-vector inputs. It measures how likely a comment will make people leave a discussion. The \texttt{attack\_on\_commenter} model measures the probability a comment is an attack on a fellow commenter and is trained on a New York Times dataset tagged by their moderation team. After inspecting the \texttt{toxicity} and \texttt{attack\_on\_commenter} scores for the tweets filtered based on the Hatebase phrases, we found that a threshold of 0.8 for \texttt{toxicity} scores and 0.5 for \texttt{attack\_on\_commenter} scores yielded a high quality dataset. 

Furthermore, to ensure directed hate speech instances attacked a specific Twitter user, we retained only those tweets that both mention another account (@) and contain second person pronouns (e.g., ``you'', ``your'', ``u'', ``ur''). The use of second person pronouns has been found to occur with high prevalence in directed hostile messages~\cite{spertus1997smokey}. The result of applying these filters is a high precision hate speech dataset of 28,318 tweets in which hate instigators use explicit Hatebase expressions against hate target accounts. 
 
\textbf{(2) Hashtag-based dataset:} 
In addition to keyphrases, we also incorporated hashtags. We examined a set of hashtags that are used heavily in the context of hate speech. We started with 13 hashtags that are likely to result in hate speech such as \#killallniggers, \#internationaloffendafeministday, \#getbackinkitchen. As we filtered the 1\% sample of Twitter's public stream from January 1st, 2016 to July 31st, 2017 for these hashtags; we eliminated hashtags with no significant presence. We include in our datasets the four hashtags that had the most hateful usage by Twitter users: \#istandwithhatespeech, \#whitepower, \#blackpeoplesuck, \#nomuslimrefugees. Finally, we obtained 597 tweets for \#istandwithhatespeech, 195 for \#whitepower, 25 for \#blackpeoplesuck, and 70 for \#nomuslimrefugees.
We include \#istandwithhatespeech in our lexical analysis but omit it from subsequent analyses because while these tweets  discuss hate speech, they are not actually hate speech themselves.



\textbf{(3) Public datasets:} To expand our hate speech corpus, we evaluate publicly available hate speech datasets and add tweet content from these datasets into our keyphrase and hashtag datasets, as appropriate. 
We start with datasets obtained by Waseem and Hovy~\cite{waseem2016hateful} and Davidson~\textit{et al.}~\cite{davidson2017automated}. We examine these existing datasets  and eliminate tweets that contain foul and offensive language but that do not fit our definition of hate speech (for example, ~\emph{``RT @usr: I can't even sit down and watch a period of women's hockey let alone a 3 hour class on it...\#notsexist just not exciting''}). We then inspect the remaining tweets and assign each  to its most appropriate hate speech category using a combination of our Hatebase keyword filter and manual annotations. Tweets that were not filtered by our Hatebase keyword approach were carefully examined and annotated manually. We obtain a total of $1,651$ tweets from~\cite{waseem2016hateful} and $1,327$ tweets from~\cite{davidson2017automated}.

Finally, we also examine hate speech reports on the No Hate Speech Movement (NHSM) website\footnote{No Hate Speech Movement Campaign: https://www.nohatespeechmovement.org/}. 
The campaign allows online users to contribute instances of hate speech on different social media platforms. We retrieve a total of $41$ English hate tweets. 

\textbf{(4) General dataset (Gen-1\%):} 
To provide a larger context for interpretation of our analyses, we compare data from our collection of hate speech datasets with a random sample of all general Twitter tweets. To create this dataset, we use the Twitter Streaming API to obtain a 1\% sample of tweets within the same 18 month collection window. From this random 1\% sample, we randomly select 85,000 English tweets.






\vspace*{0.05in}
\noindent
{\bf Human-centered dataset evaluation.}
We evaluate the quality of our final datasets by incorporating human judgment using Crowdflower. We provided annotators with a class balanced random sample of $2000$ tweets and asked them to annotate whether or not the tweet was hate speech or not, and whether the tweet was directed towards a group of people (Generalized hate speech) or directed towards an individual (Directed hate speech). To aid annotation, all annotators were provided a set of precise instructions. This included the definition of hate speech according to the social media community (Facebook and Twitter) and examples of hate tweets selected from each of our eight hate speech categories. Each tweet was labeled by at least three independent Crowdflower annotators, and all annotators were required to maintain at least an 80\% accuracy based on their performance of five test questions - falling below this accuracy resulted in automatic removal from the task. We then measured the inter-annotator reliability to assess the quality of our dataset. For the representative sample from our Generalized hate speech dataset, annotators labeled 95.6\% of the tweets as hate speech and 87.5\% of tweets as hate speech directed towards a group of people. For the representative sample from our Directed hate speech dataset, annotators labeled 97.8\% of the tweets as hate speech and 94.3\% of tweets as hate speech directed towards an individual. Our dataset obtained a Krippendorf's alpha of 0.622, which is 38\% higher than other crowd-sourced studies that observed online harmful behavior~\cite{wulczyn2017ex}.

\subsection{Measures}
In our investigation, we adopt several measures based on prior work in order to study linguistic features that differentiate between Directed and Generalized hate speech. To alleviate the effects of domain shift in our choice of models, we use tools that are developed and trained using Twitter data when available and fall back to state of the art models that were trained on English data in the event of unavailability of Twitter-specific tools. To analyze the salient words for each category of hate speech keywords (e.g., ethnicity, class, gender) and specific language semantics associated with hashtags, we use SAGE~\cite{eisenstein2011sparse}, a mixed-effect topic model that implements the L1-regularized version of sparse additive generative models of text. SAGE has been used in several Natural Language Processing (NLP) applications including~\cite{sim2012discovering} that provides a joint probabilistic model of who cites whom in computational linguistics, and~\cite{wang2012historical} which aims to understand how opinions change temporally around the topic of slavery-related United States property law judgments. To extract entities from the collected tweets, we leverage T-NER, a system developed specifically to perform the task of Named Entity Recognition on tweets~\cite{Ritter11}. To understand the linguistic dimension and psychological processes identified among Directed hate, Generalized hate, and general Twitter tweets, we use the psycholinguistic lexicon software LIWC2015~\cite{chung-pennebaker}, a text analysis tool that measures psychological dimensions, such as affection and cognition. To analyze frame semantics of hate speech, we use \textsc{SemaFor} \cite{chen2010semafor}, which annotates text with their evoked frames as defined by \textsc{FrameNet} \cite{baker1998berkeley,ruppenhofer2006framenet}. While we acknowledge that \textsc{Semafor} is not trained on Twitter, it has been found that it is actually more robust to domain-shift and its performance on Twitter is comparable to that on Newswire~\cite{sogaard2015using}.   

\section{Analysis}
\begin{table}[!tb]
\small
\centering
\resizebox{\columnwidth}{!}{%

\begin{tabular}{c|c||c|c}
\hline
 \textbf{Archaic Generalized} & \textbf{Archaic Directed} &  \textbf{Class Generalized} &  \textbf{Class Directed} \\ \hline
Anti & hillbilly  & Catholics & Rube \\
wigger & chinaman & hollering & \#redneck\\
hillbilly & verbally & \#racist & ALABAMA \\
bitch & prostitute & Cracker & batshit \\
white  & vegetables & \#Virginia & DRINKS \\ \hline \hline
 
\textbf{Disability Generalized} &  \textbf{Disability Directed} &  \textbf{Ethnicity Generalized} & \textbf{Ethnicity Directed} \\ \hline
 retards & \#Retard   & Anglo &  coons \\
 legit & sniping &  spics & Redskins \\
 Only & \#retarded  & breeds & Rhodes\\
 yo &  Asshole  & hollering & \#wifebeater\\
 phone &  upbringing & actin & plantation\\ \hline \hline

 \textbf{Gender Generalized} &  \textbf{Gender Directed} &  \textbf{Nationality Generalized} & \textbf{Nationality Directed} \\ \hline
 dyke(s) & \#CUNT  & Anti& chinaman\\
 chick & judgemental & wigger& Zionazi(s)\\
 cunts & aitercation & bitch & \#BoycottIsrael\\
 hoes & Scouse & white & prostitute\\
 bitches & traitorous & & \#BDS\\ \hline \hline

 \textbf{Religion Generalized} &  \textbf{Religion Directed} &  \textbf{SexOrient Generalized} & \textbf{SexOrient Directed} \\ \hline
 Algebra & catapults & meh& pansy \\
 Israelis & Muzzie & \#faggot(s)&  Cuck\\
 extermination & Zionazi & queers &  CHILDREN\\
 Jihadi & \#BoycottIsrael & hipster & FOH\\
 lunatics &  rationalize & NFL & wrists\\

 \hline

\hline
\end{tabular}
}
\caption{Top five keywords learned by SAGE for each hate speech class. Note the presence of distinctive words related to each class (both for Generalized and Directed hate).}
\label{table:SAGE_hatetype}
\end{table}

\subsection{Lexical Analysis}
To analyze salient words that characterize different hate speech types, we use SAGE~\cite{eisenstein2011sparse}. SAGE offers the advantages of being supervised, building relatively clean topic models by taking into account additive effects and combining multiple generative facets, including background, topic and perspective distributions of words.
In our analysis, each tweet is treated as a document and we only include words that appear at least five times in the entire corpus. This step is crucial to ensure that SAGE's supervised learning model will find salient words that  not only identify each hate speech type or hashtag, but also are well-represented in our datasets. 

\vspace*{0.05in}
\noindent
\textbf{What are the salient words characterizing different hate speech categories?} 
Table~\ref{table:SAGE_hatetype} shows the top five salient words learned by SAGE for each hate speech type. We note that there is minimal intersection of salient words between different hate speech categories, e.g., ethnicity, archaic, and SexOrient, and between the generalized and directed versions of each hate speech type. Although a tweet could contain several keywords pertaining to different types of hate speech, the top salient words indicate that hate speech categories have distinct topic domains with minimal overlap. For example, note the presence of words \texttt{retards, \#Retard} used in hate speech related to disability. Similarly, note the presence of religion related words like \texttt{Jihadis, extermination, Zionazi, Muzzie} for religion-related hate speech.

\begin{figure}[!tb]
   \centering
        \subfloat[\#whitepower]{\includegraphics[width=0.4\textwidth, height = 3.5cm]{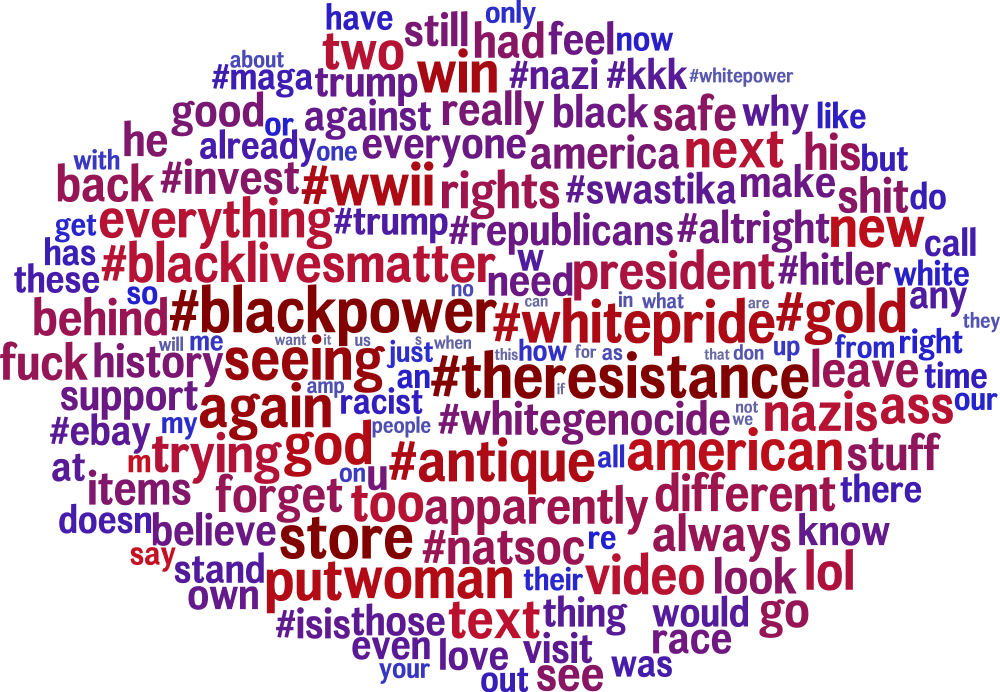}\label{fig:whitepower}} \\
       \subfloat[\#nomuslimrefugees] {\includegraphics[width=0.4\textwidth, height = 3.5cm]{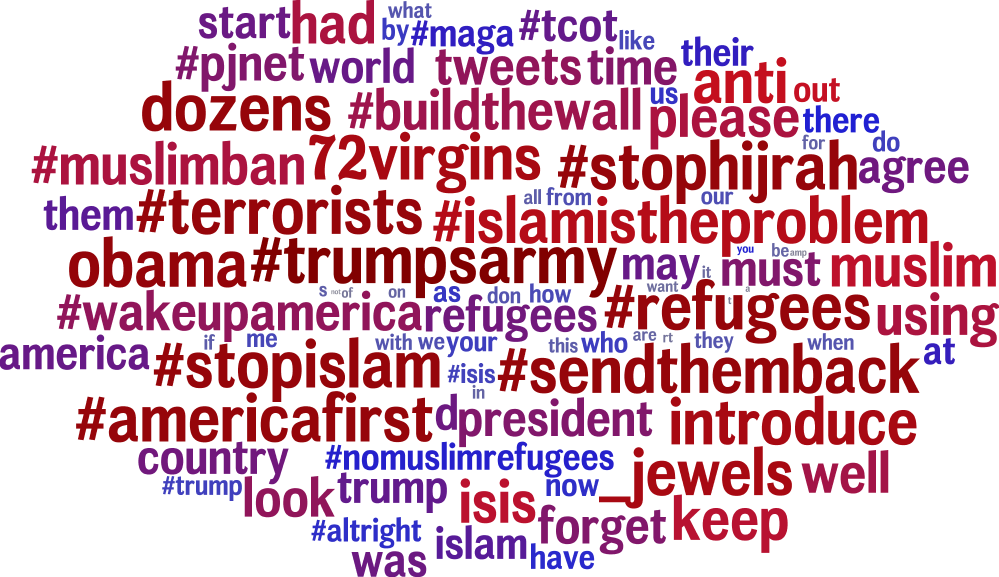}\label{fig:nomuslimrefugees}}
   \caption{The salient words for tweets associated with \#whitepower and \#nomuslimrefugees learned by the  sparse additive generative model of text. A larger font corresponds to a higher score output by the model.}
 \label{fig:hashtags-SAGE}
\end{figure}

We show the results of SAGE for the hashtags \#whitepower (categorized as ethnicity-based hate) and \#nomuslimrefugees (categorized as religion-based hate) in Figure~\ref{fig:hashtags-SAGE}. Among the salient words for the hashtag \#whitepower are \#whitepride, \#whitegenocide, the resistance, \#wwii, nazi, \#kkk, \#altright, and republicans. For the hashtag \#nomuslimrefugees, salient words include \#stopislam, \#islamistheproblem, \#trumpsarmy, \#terrorists, \#muslimban, \#sendthemback, and \#americafirst. 

\vspace*{0.05in}
\noindent \textbf{What are the prevalent themes in hate speech participation?} We examine the salient words for \#istandwithhatespeech to gain  insight into why people participate in hate speech. The top five salient words for  \#istandwithhatespeech are \textit{banned, allowed, opinion, \#1a,} and~\textit{violence}. Further inspection of tweets for these keywords revealed the following themes: \textbf{(a)} hate and other  offensive speech should be allowed on the Internet; \textbf{(b)} not participating in hate speech implies the inability to handle different opinions; \textbf{(c)} hate speech is truth telling; and \textbf{(d)} the First Amendment (\#1a) grants the right to participate in hate speech. Some example tweets representing these viewpoints include: \textit{@usr: people should be allowed to tell the truth no matter how it affects other people. \#istandwithhatespeech};  \textit{@usr: \#istandwithhatespeech because the eu shouldn't dictate what is allowed on the internet, a global communication system}; and \textit{\#istandwithhatespeech b/c if you really can't hear an opinion different from your own you need f*cking therapy.}


\vspace*{0.05in}
\noindent 
\textbf{How are named entities represented across Directed and Generalized hate?}
Named Entity Recognition seeks to identify names of persons, organizations, locations, expressions of times, brands, and companies among other categories within selected text. For example, consider the following tweet: \textit{``@usr Obama and Hillary ain't gone protect you when trump is president. btw you need some braces you f*ckin dyke.''} The task of Named Entity Recognition would  identify~\textit{Obama, Hillary}, and~\textit{trump} as person entities.

\begin{figure}[!tb]
   \centering
       {\includegraphics[width=0.8\linewidth]{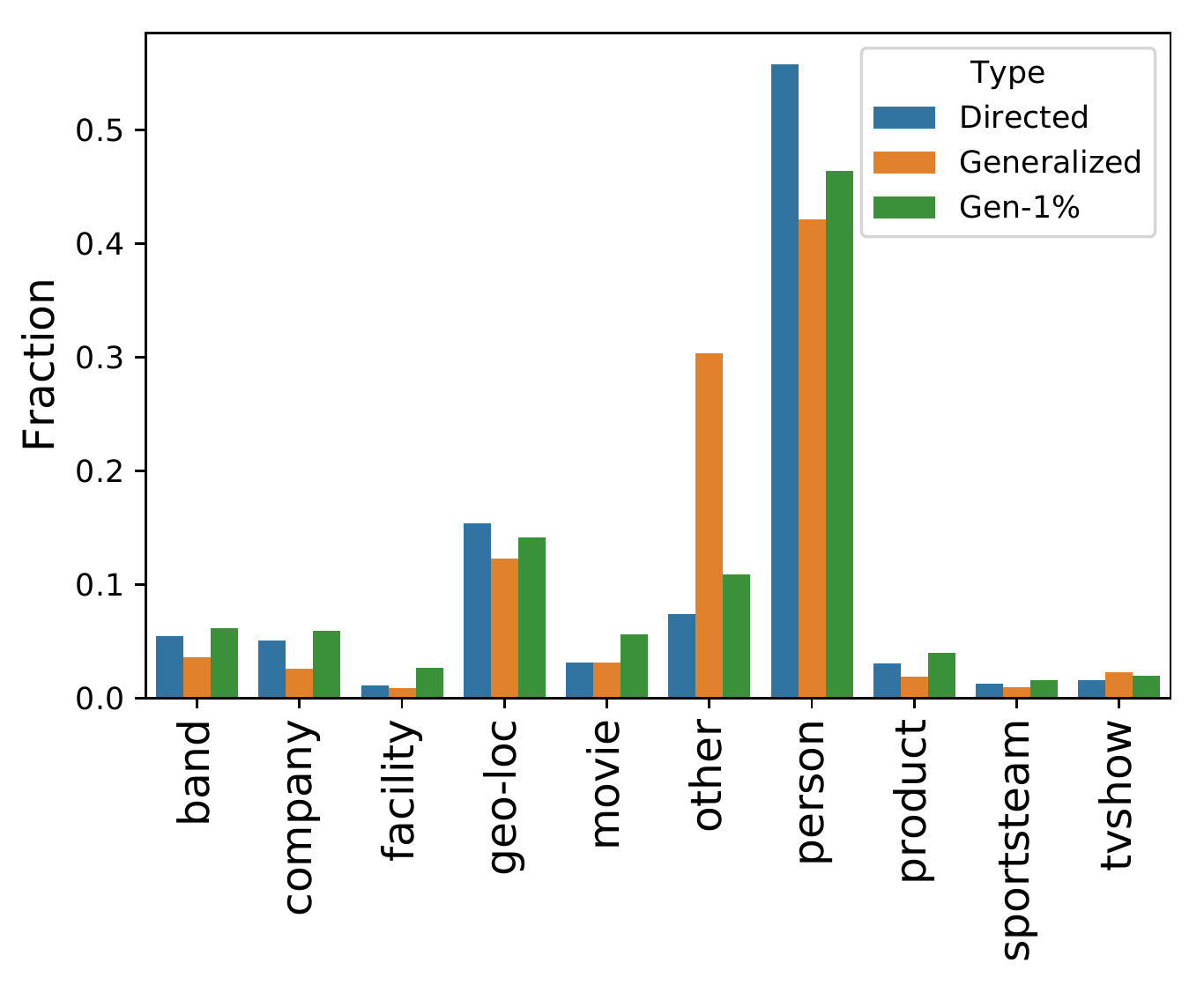}\label{fig:summary_entities}}
        \caption{Proportion of entity types in hate speech. Note the much higher proportion of \textsc{Person} mentions in Directed hate speech, suggesting direct attacks. In contrast, there is a higher proportion of \textsc{Other} in Generalized hate speech, which are primarily religious entities (i.e. \texttt{Islam, Muslim, Jews, Christians}).}
 \label{fig:entities_barchart}
\end{figure}

Figure~\ref{fig:entities_barchart} shows a breakdown of entities identified by T-NER for Directed hate, Generalized hate and Gen-1\% tweets. We first note that Directed hate tends to have a higher percentage of person entities (55.8\%) as opposed to Generalized hate (42.1\%), and Gen-1\% (46.4\%). This is expected since Directed hate speech is often a personal attack on specific person(s). We find that tweets have other entities that do not belong to persons, brands, companies, facilities, geo-locations, movies, products, sports teams or TV shows.  These include Islam and Jews; we separate these tweets into an ``other'' category. 

\begin{figure*}[!htb]
   \centering
        \subfloat[Directed hate]{\includegraphics[width=0.33\linewidth]{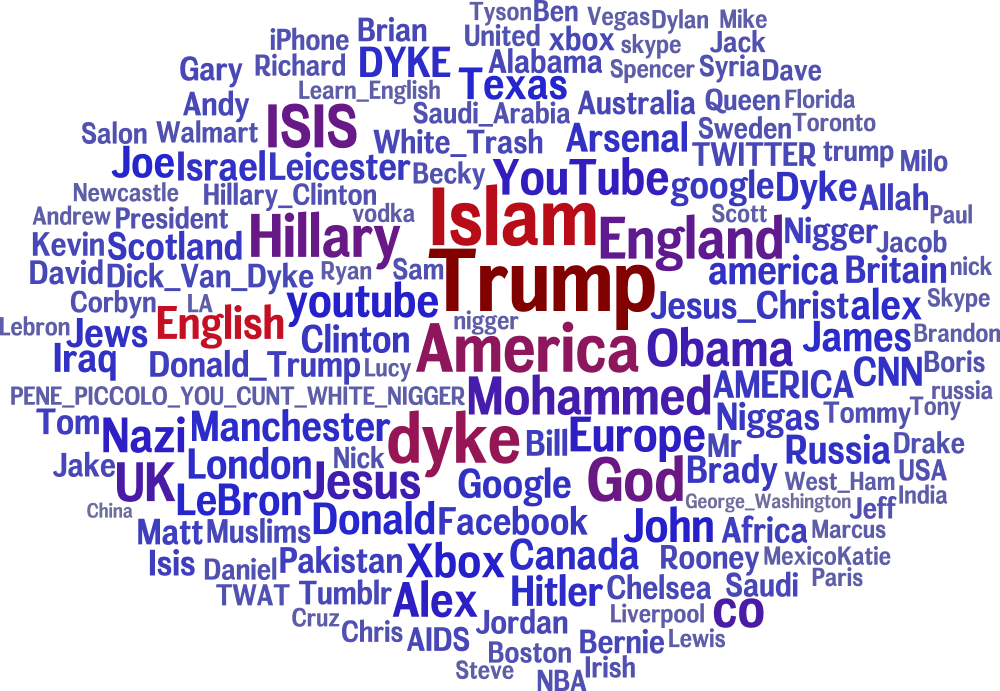}\label{fig:direct_entities}}
       \subfloat[Generalized hate] {\includegraphics[width=0.33\linewidth]{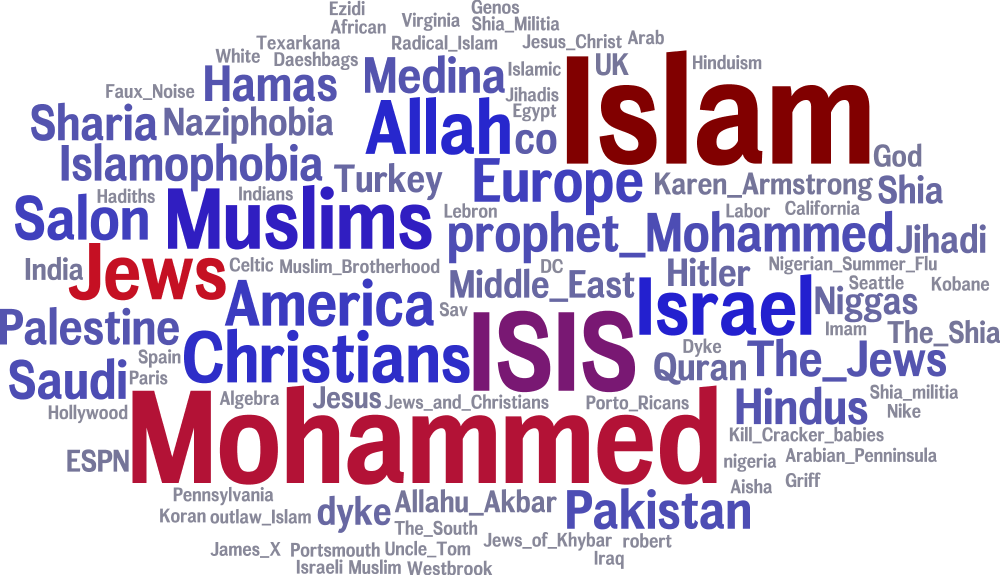}\label{fig:generalized_entities}}
        \subfloat[General-1\%] {\includegraphics[width=0.33\linewidth]{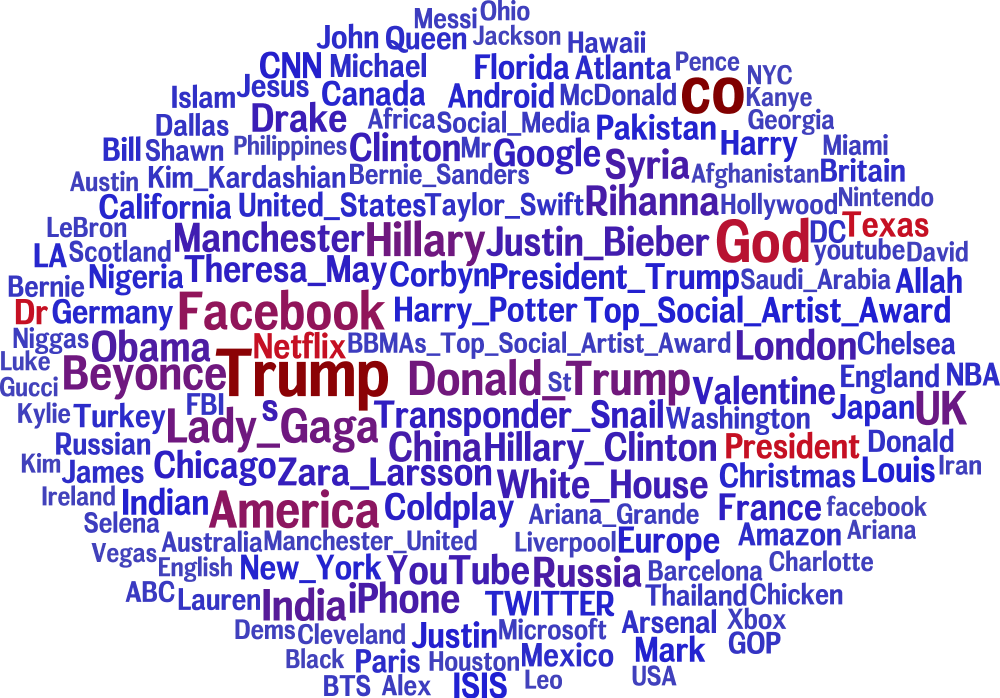}\label{fig:one_entities}} \\
       \caption{Top entity mentions in Directed, Generalized and Gen-1\% sample. Note the presence of many more person names in Directed hate speech. Generalized hate speech is  dominated by religious and ethnicity words, while  the random $1\%$ is dominated by celebrity names.}
 \label{fig:entities_wordcloud}
\end{figure*}

We inspect all the entities recognized by T-NER and represent them in Figure~\ref{fig:entities_wordcloud}. We note that some entities are universally present in different categories. These include~\textit{Trump, Hillary, Islam, Mohammed, Google, ISIS,} and \textit{America}. Additionally, we find that Directed hate contains more common names such as~\textit{Scott, Sam, Andrew, Katie, Ben, Ryan, Jamie,} and \textit{Lucy}. Generalized hate tends to contain religious-based entities such as~\textit{Jews, Muslims, Christians, Hindus, Shia, Madina,} and \textit{Hammas}, and entities involved in political and religious disputes and conflicts such as~\textit{Hamas, Palestine,} and \textit{Israel}. This is also consistent with our observation that the majority of the Generalized hate speech tweets happen to be related to \textsc{Religion} (although no specific filtering for religion was done in the data collection step). On the other hand, we observe that certain popular individuals, such as~\textit{Theresa May, Beyonce, Justin Bieber, Lady Gaga, Taylor Swift, Tom Brady,} and \textit{Katy Perry}, exist only in Gen-1\%, suggesting that these  categories differ in their focus.

In summary, our lexical analysis highlights salient features and entities that distinguish between Directed and Generalized hate speech while also revealing evident themes that indicate why people choose to participate in hate speech. 



\subsection{Psycholinguistic Analysis}



\begin{figure*}[!tb]
   \centering
        \subfloat[Summary]{\includegraphics[width=0.33\linewidth]{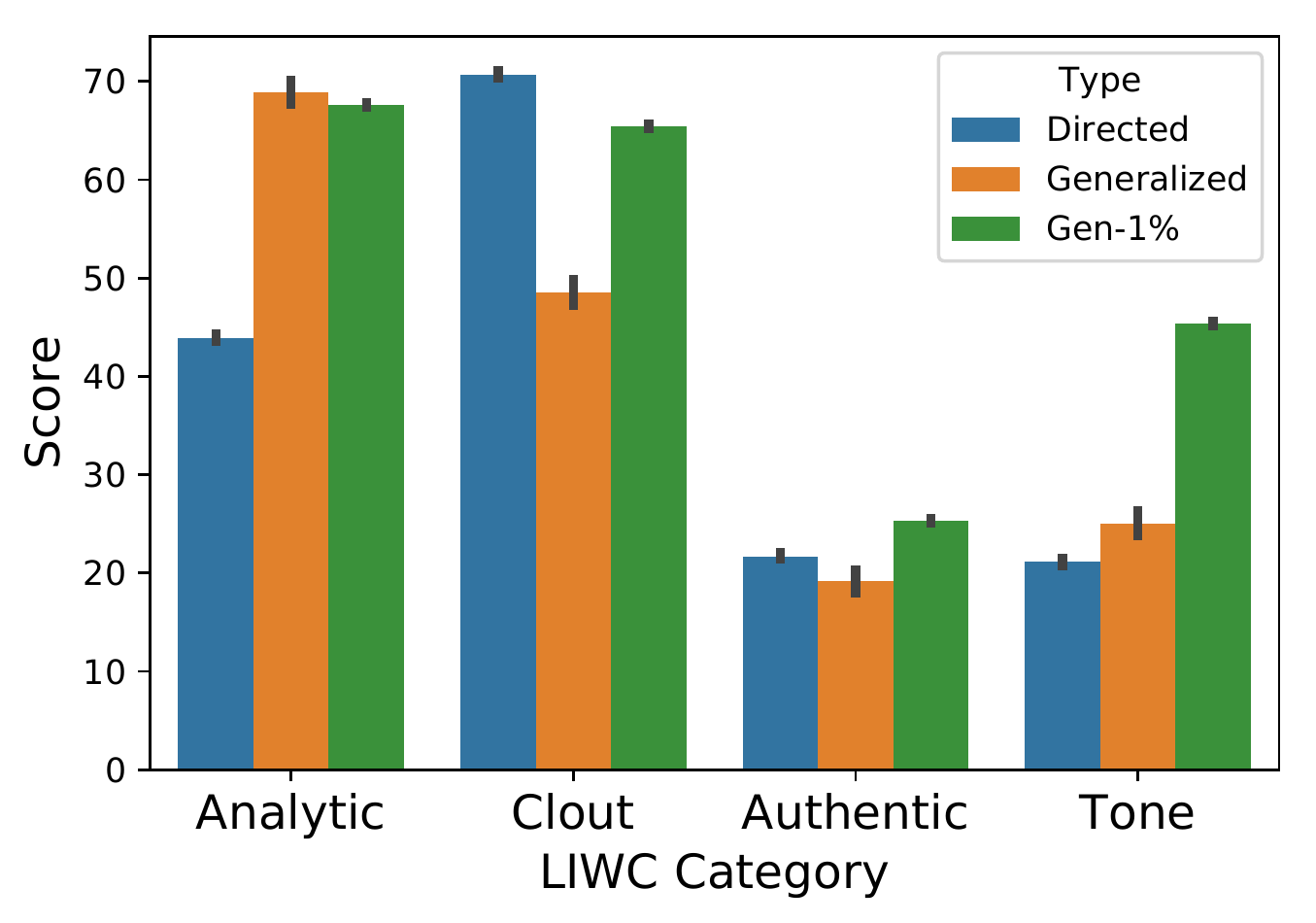}\label{fig:liwc_summary}}
       \subfloat[Psychological processes] {\includegraphics[width=0.33\linewidth]{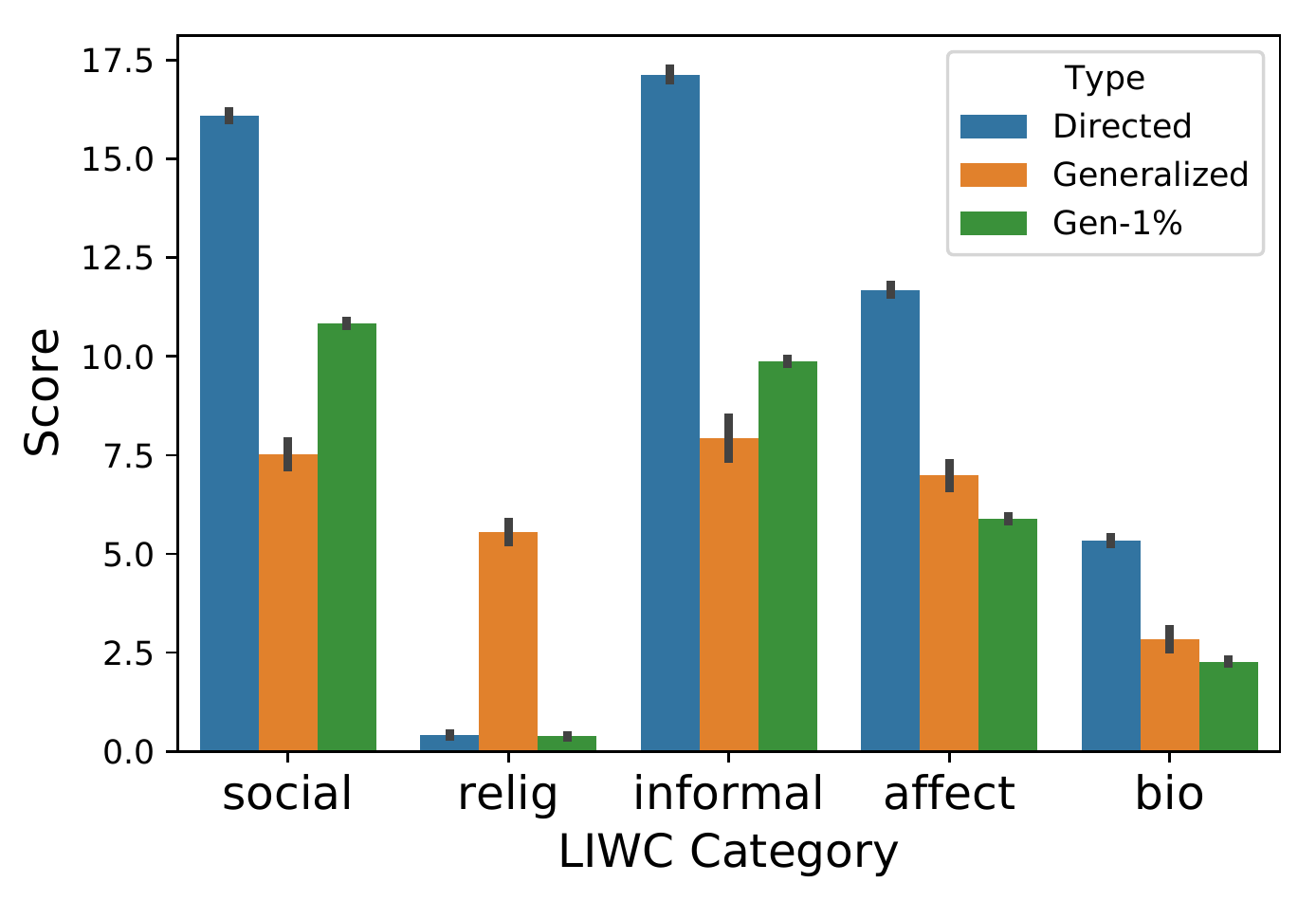}\label{fig:liwc_psych}}
        \subfloat[Person pronouns] {\includegraphics[width=0.33\linewidth]{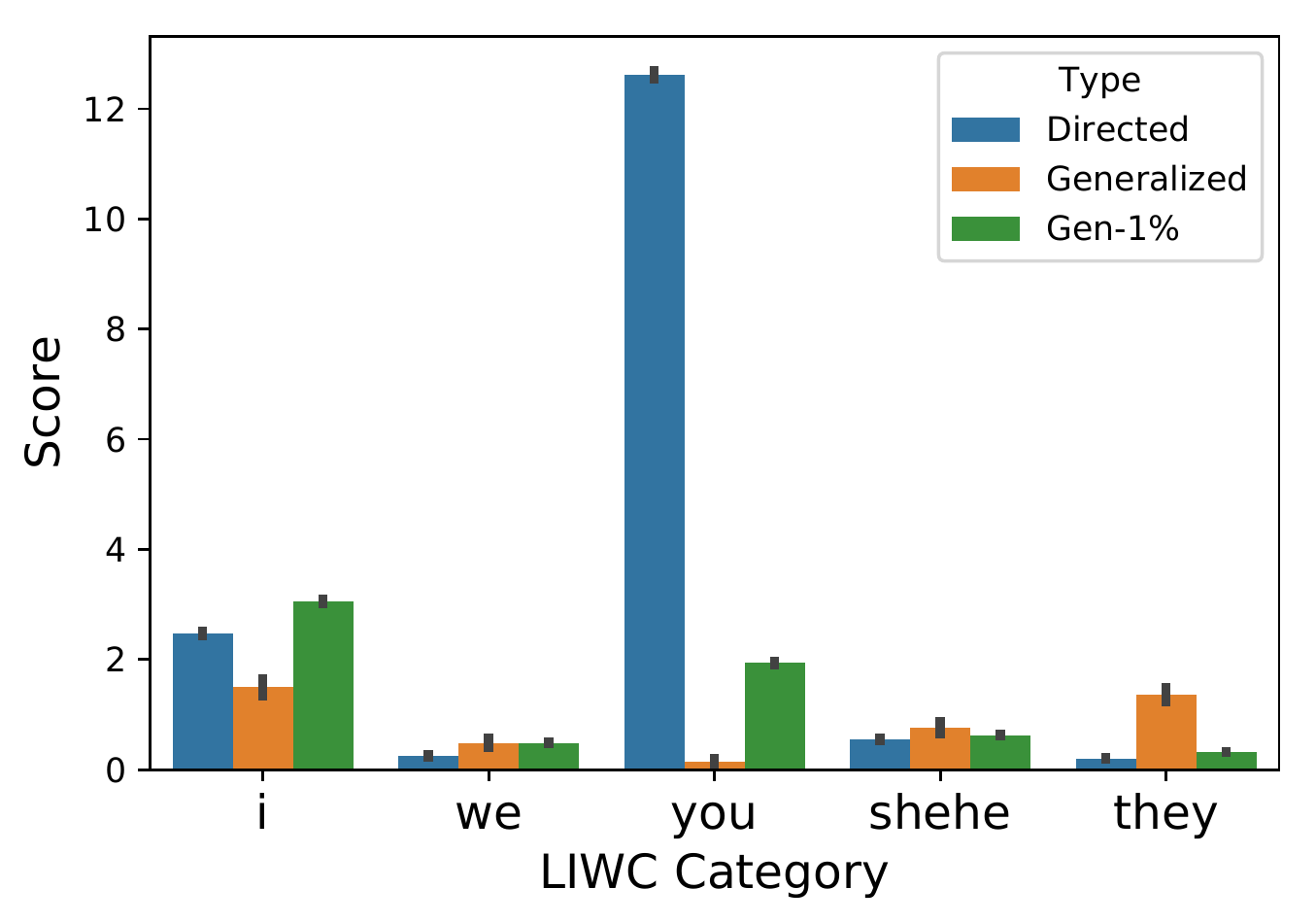}\label{fig:targets-hist-log}} \\
       \subfloat[Negative emotions]{\includegraphics[width=0.33\linewidth]{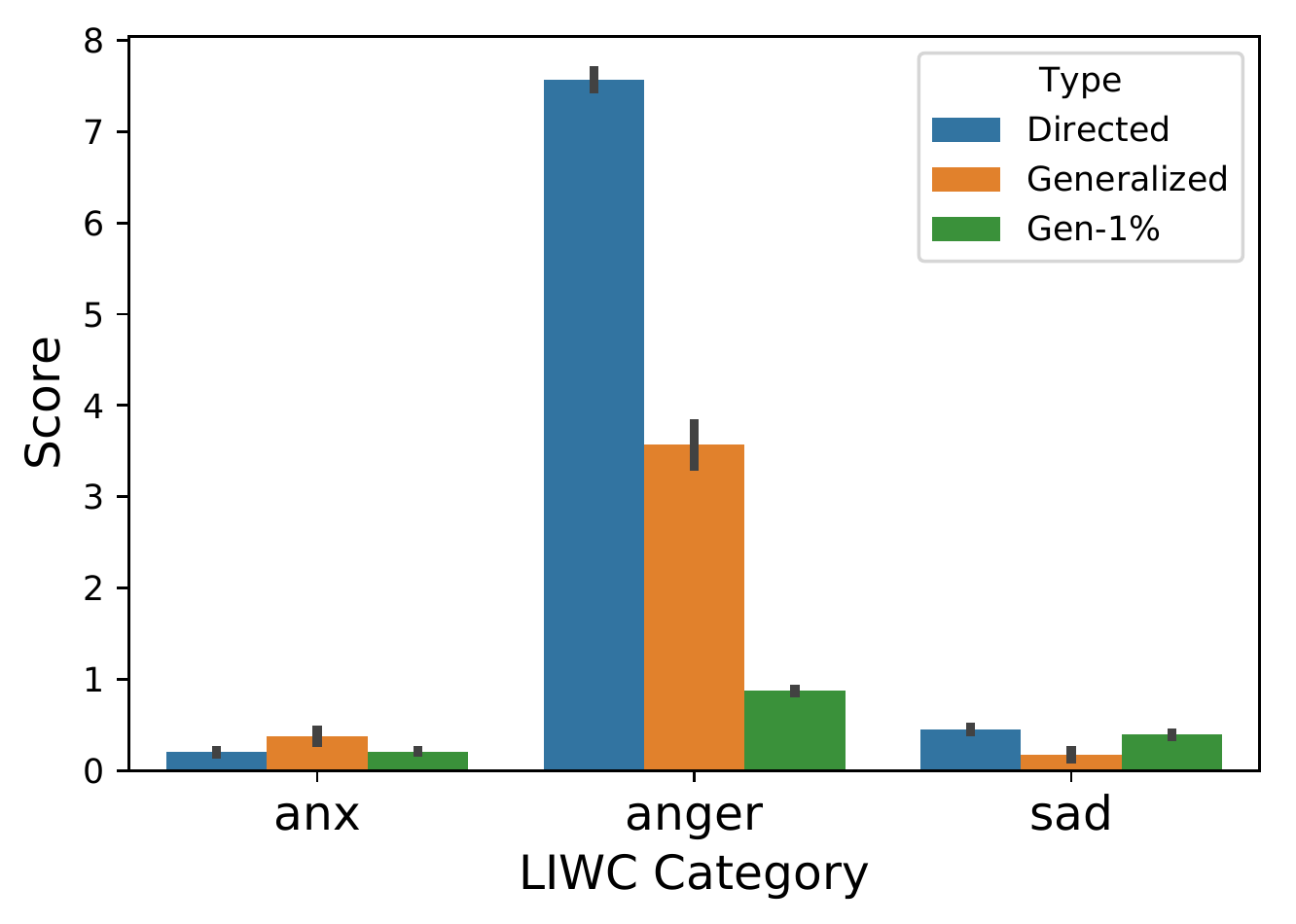}\label{fig:anx}}
       \subfloat[Temporal focus] {\includegraphics[width=0.33\linewidth]{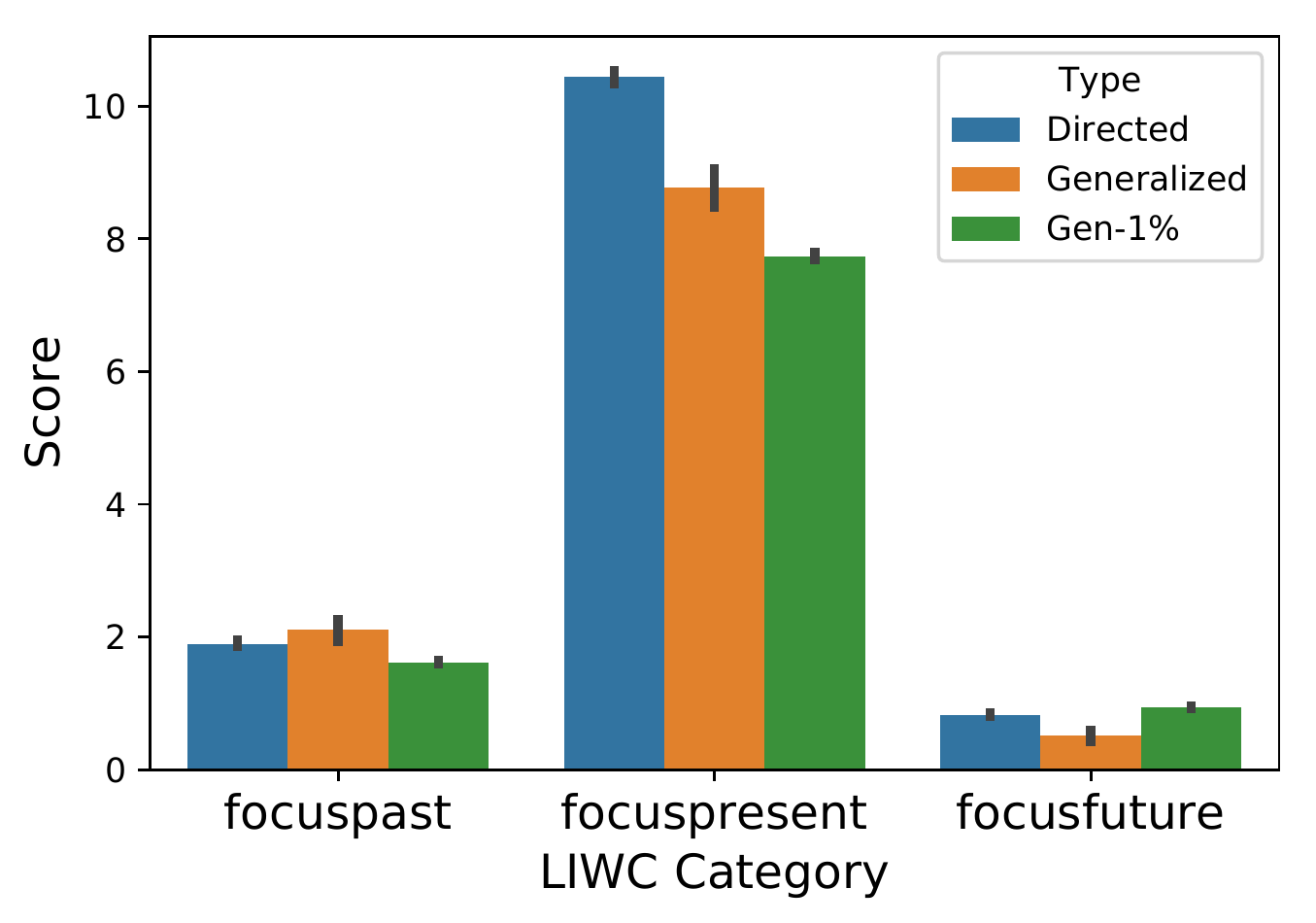}\label{fig:focus}}
        \subfloat[Personal concerns] {\includegraphics[width=0.33\linewidth]{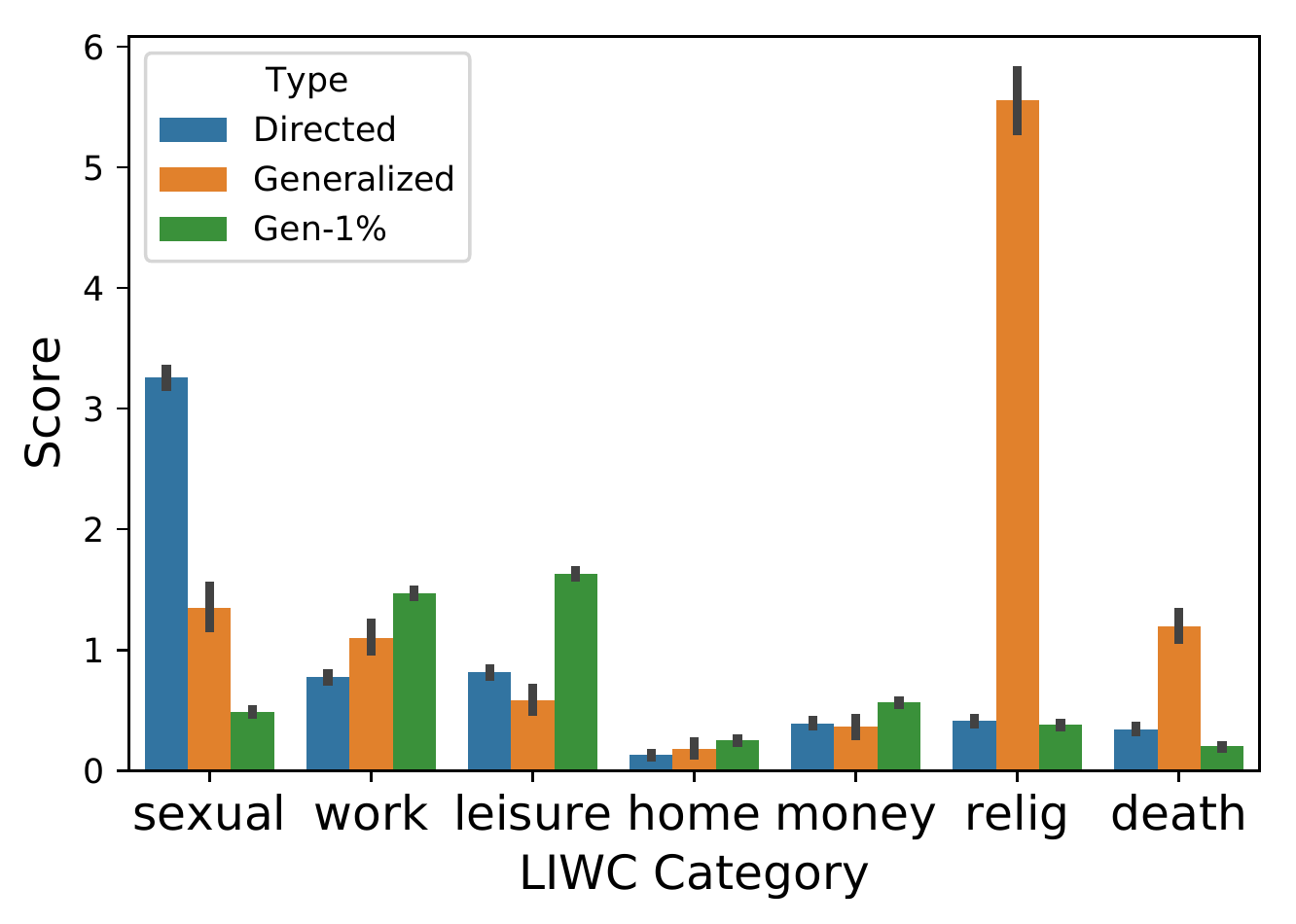}\label{fig:personal}} \\
       \caption{Mean scores for LIWC categories. Several differences exist between Directed hate speech and Generalized hate speech. For example,  Directed hate speech exhibits  more anger than Generalized hate speech, and Generalized hate speech is primarily associated with religion. Error bars show 95\% confidence intervals of the mean.}
 \label{fig:liwc_results}
\end{figure*}
For a full psycho-linguistic analysis, we use LIWC~\cite{chung-pennebaker}. Specifically, we focus on  the following dimensions: summary scores, psychological processes, and linguistic dimensions. A detailed description of these dimensions and their attributes can be found in the LIWC2015 language manual~\cite{chung-pennebaker}. Figure~\ref{fig:liwc_results} shows the mean scores for our key LIWC attributes. Our analysis yields the following observations.

\vspace*{0.04in}
\noindent \textbf{Directed hate speech exhibits the highest clout and the least analytical thinking, while general tweets exhibit the highest authenticity and emotional tone.}
Figure~\ref{fig:liwc_results}(a) shows the key summary language values obtained from LIWC2015 averaged over all tweets for Directed hate, Generalized hate, and Gen-1\%. We show that Directed hate has the lowest mean for analytical thinking scores ($\mu$ = 43.9, $p < 0.001$) in comparison to Generalized hate ($\mu$ = 68.9) and Gen-1\% ($\mu$ = 67.6). We also note that Directed hate demonstrates higher mean clout (influence and power) values ($\mu$ = 70.7, $p < 0.001$) than Generalized hate ($\mu$ = 48.5) and Gen-1\% ($\mu$ = 65.4). This result resonates with the nature of personal directed hate attacks, in which persons exhibit dominance and power over others. Moreover, Figure~\ref{fig:liwc_results}~(a) indicates that tweets in the Gen-1\% dataset have the highest mean value of authenticity (Authentic) ($\mu$ = 25.3, $p < 0.001$) in comparison to hate tweets: directed ($\mu$ = 21.7) and generalized ($\mu$ = 19.2). Additionally, we note that Gen-1\% ($\mu$ = 41.4, $p < 0.001$) has the highest mean score of emotional tone (Tone) followed by Generalized ($\mu$ = 25.1) and Directed hate ($\mu$ = 21.1). This indicates that general tweets are associated with a more positive tone, while Generalized and Directed hate language reveal greater hostility.

\vspace*{0.03in}
\noindent \textbf{Directed hate speech is  more informal and social than generalized hate and general tweets.}
Figure~\ref{fig:liwc_results}(b) shows that Directed hate has a much higher mean informal score ($\mu$ = 17.1, $p < 0.001$) in comparison to generalized hate ($\mu$ = 7.9) and Gen-1\% ($\mu$ = 9.9). Informality includes the usage of swear words and abbreviations, e.g., btw, thx. Additionally, Directed hate tends to have higher social components ($\mu$ = 16.1 vs. 7.5 for generalized hate and 10.9 for general tweets, $p < 0.001$) inherent in its linguistic style, which manifests in greater usage of language related to family, friends, and male and female references.

\vspace*{0.03in}
\noindent \textbf{Generalized hate speech emphasizes ``they'' and not ``we''.}
Figure~\ref{fig:liwc_results}(c) shows that generalized hate speech has higher usage of third personal plural pronouns (they) than first personal plural pronouns (we). The mean score for third person pronoun usage is 1.4, in comparison to 0.5; 2.8x higher ($p < 0.001$). An example tweet is: ``\textit{Muslims are not a race, idiot, they are a cult of murder and terrorism.}''

\vspace*{0.03in}
\noindent \textbf{Directed hate speech is angrier than generalized hate speech, which in turn is angrier than general tweets.} We show that anger manifests differently across Generalized and Directed hate speech. Figure~\ref{fig:liwc_results}(d) shows that Directed hate contains the angriest voices ($\mu$ = 7.6, $p < 0.001$) followed by Generalized hate ($\mu$ = 3.6); general tweets are the least angry ($\mu$ = 0.9). In~\cite{cheng2017anyone}, the authors observe that negative mood increased a user's probability to engage in trolling, and that anger begets more anger. Our results complement this observation by differentiating between levels of anger for Directed and Generalized hate. Example tweets include: ``\textit{@usr F*ckin muzzie c*nts, should all be deported, savages}'' and ``\textit{f*ck n*ggers, faggots, chinks, sand n*ggers and everyone who isnt white}.'' 

\vspace*{0.03in}
\noindent \textbf{Both categories of hate speech are more focused on the present than general tweets.}
Figure~\ref{fig:liwc_results}(e) shows that hate speech ($\mu$ = 10.4 and = 8.7 for Directed and Generalized hate, respectively, $p < 0.001$) more commonly emphasizes the  present than general tweets ($\mu$ = 7.7). Examples include: ``\textit{How the f*ck does a foreigner win miss America? She is Arab! \#idiots}'' and ``\textit{@usr Those n*ggers disgust me. They should have dealt with 100 years ago, we wouldn't be having these problems now}''.


\newpage
\noindent \textbf{General tweets have the fewest sexual references while generalized hate has the most death references.}
Figure~\ref{fig:liwc_results}(e) shows that general tweets have the lowest mean score for sexual references ($\mu$ = 0.5, $p < 0.001$) in comparison to Directed hate ($\mu$ = 3.3) and Generalized hate ($\mu$ = 1.3). Moreover, our analysis shows that, compared to general tweets ($\mu$ = 0.2), hate tweets are more likely to incorporate death language ($\mu$ = 1.2, $p = 0.1$ for Generalized hate and = 0.34 for Directed hate, $p < 0.001$).

\subsection{Semantic Analysis}

\begin{figure}[]
   \centering
        {\includegraphics[width=0.9\linewidth]{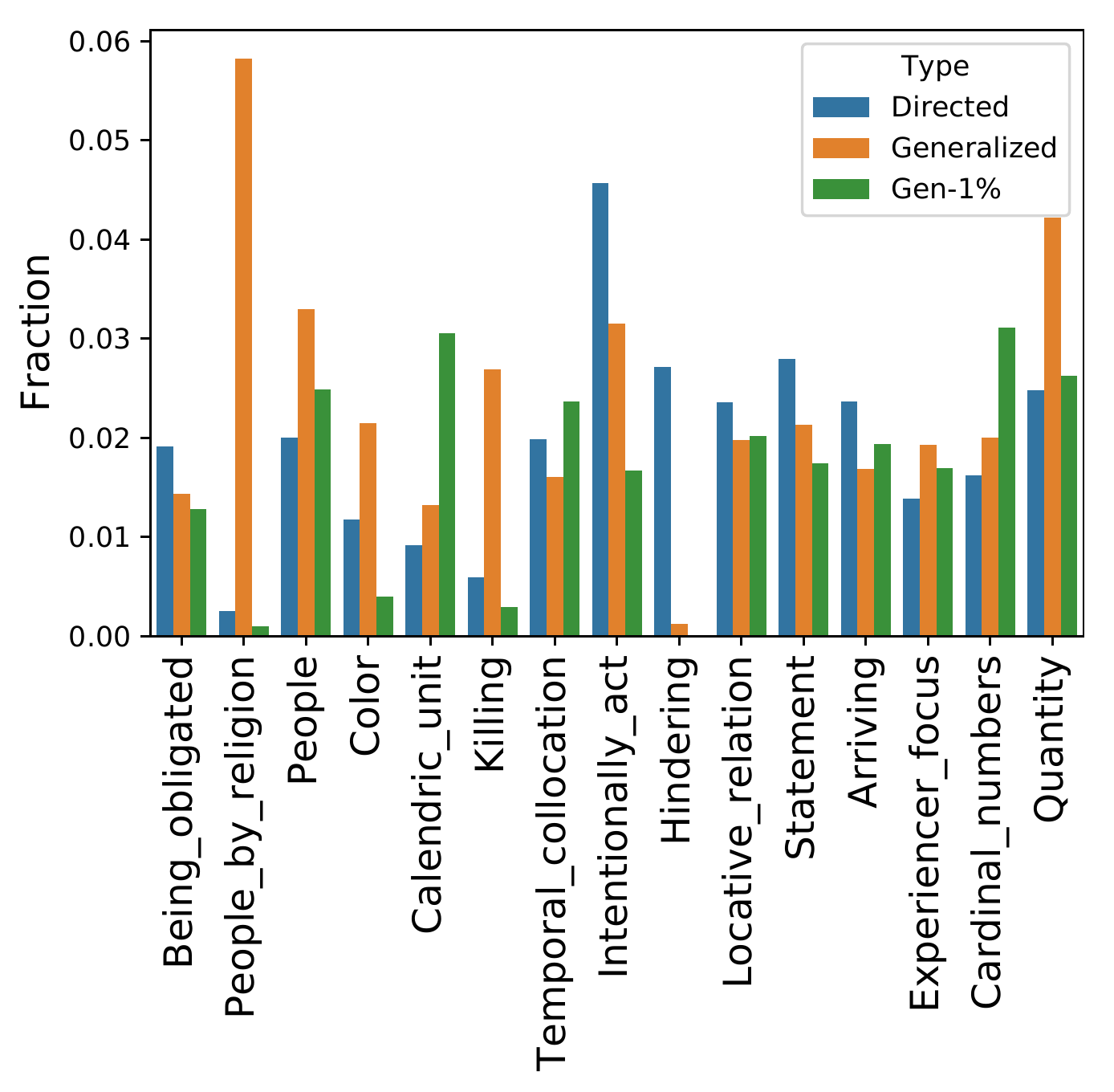}\label{fig:semafor_summary}}
        \caption{Proportion of frames in different types. Note the much higher proportion of \textsc{People\_by\_religion} frame mentions in Generalized hate speech. In contrast, Directed hate speech evokes frames such as \textsc{Intentionally\_act} and \textsc{Hindering}.}
 \label{fig:semafor_barchart}
\end{figure}

\begin{figure*}[!htb]
   \centering
        \subfloat[Directed hate]{\includegraphics[width=0.3\linewidth]{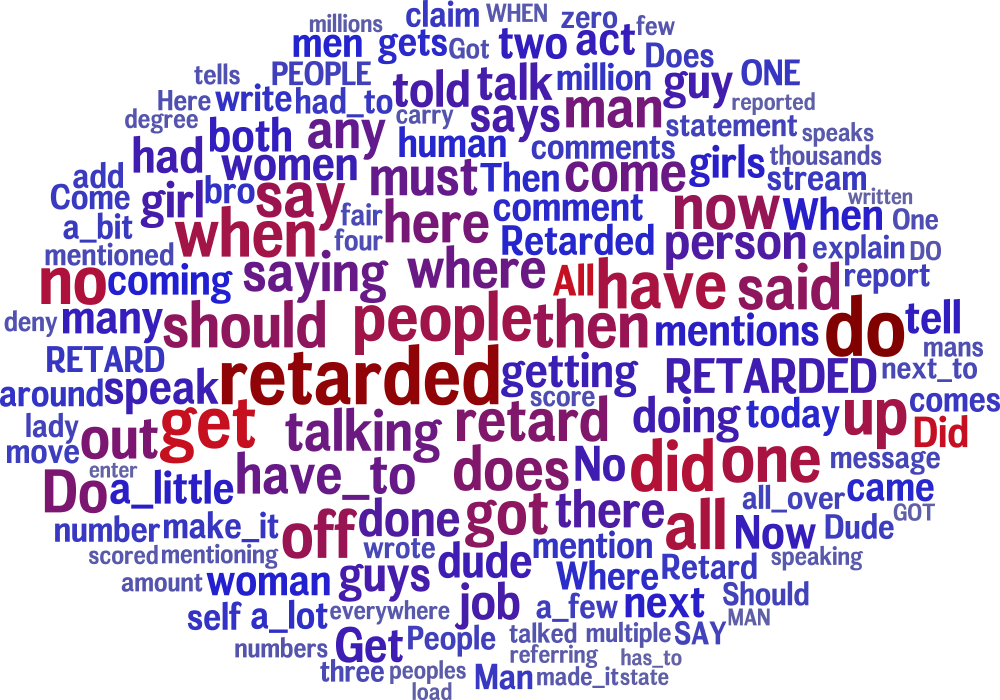}\label{fig:direct_entities}}
       \subfloat[Generalized hate] {\includegraphics[width=0.3\linewidth]{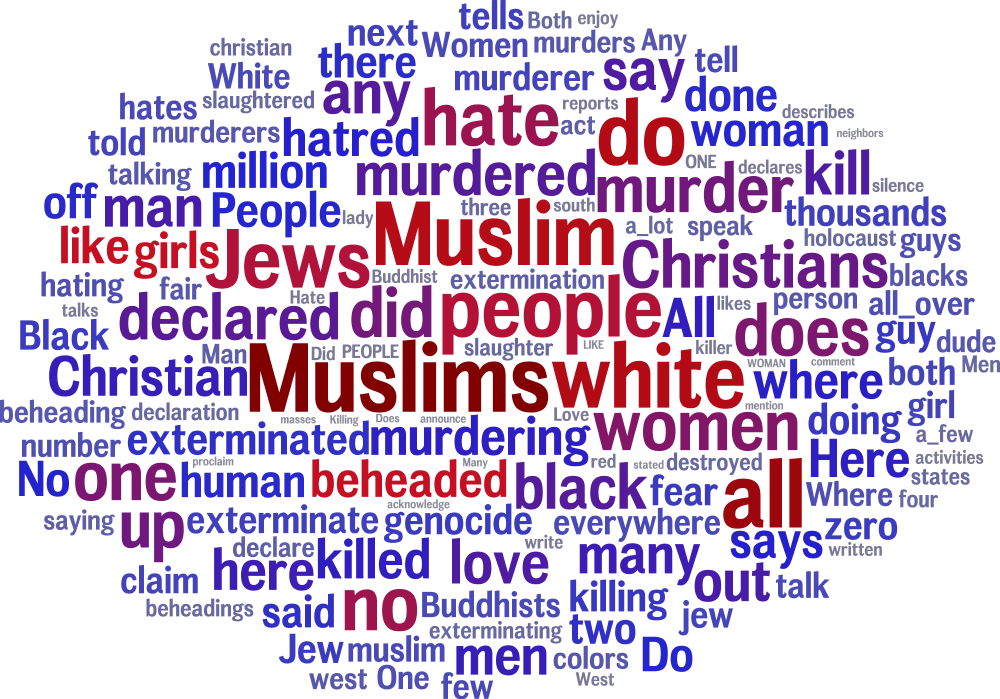}\label{fig:generalized_entities}}
       \subfloat[Gen-1\%] {\includegraphics[width=0.3\linewidth]{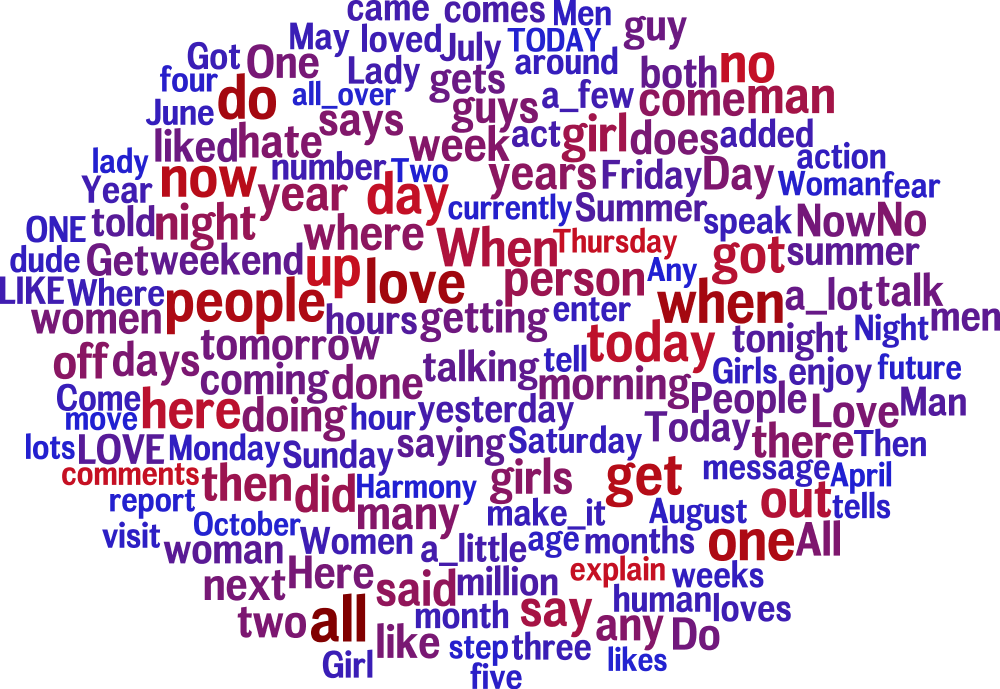}\label{fig:random_entities}}
       \caption{Words evoked by the top 10 semantic frames in each hate class. In Directed hate speech, note the presence of action words such as \texttt{do, did, now, saying, must, done} and words that condemn actions  (\texttt{retard, retarded}). In sharp contrast, Generalized hate speech evokes words related to \textsc{Killing}, \textsc{Religion} and \textsc{Quantity} such as \texttt{Muslim, Muslims, Jews, Christian, murder, killed, kill, exterminated,} and \texttt{million}.}
 \label{fig:semafor_wc}
\end{figure*}

In this section, we  turn our attention to the frame-semantics of the hate speech categories. Using frame-semantics, we can analyze higher-level rich structures called \emph{frames} that represent real world concepts (or stereotypical situations) that are evoked by words. For example, the frame \textsc{Attack} would represent the concept of a \texttt{person} being attacked by an \texttt{attacker} with perhaps a \texttt{weapon} situated at some point in \texttt{space} and \texttt{time}. 

After annotating Directed and Generalized hate speech tweets using~\textsc{SemaFor}, we compute the distribution over evoked frames for each type of hate speech. Figure~\ref{fig:semafor_barchart} shows proportions for $15$ frame types (top 5 from each type) for Directed hate, Generalized hate and Gen-1\%. We make the following observations.

\vspace*{0.05in}
\noindent
\textbf{Directed hate speech evokes \emph{intentional acts, statements} and \emph{hindering.}}
Our analysis reveals that the Directed hate speech has a higher proportion of intentionally\_act frames (0.05, $p < 0.01$) than generalized hate (0.03) and general tweets (0.016). An example of a tweet with an intentionally\_act frame is: ``\textit{@usr if you \textbf{don't}\footnote{Bold font indicates words that evoked the corresponding frames.} choose @usr you're the biggest f*ggot to ever touch the face of the earth}''.
Moreover, Directed hate has the highest proportion of statement frames and hindering frames (0.03 and 0.03, respectively, $p < 0.01$) when compared to generalized hate (0.02 and 0.001) and general tweets (0.017 and 0.0001). Examples of tweets with statement and hindering frames are: ``\textit{I do not like \textbf{talking} to you f*ggot and I did but in a nicely way f*g}'' and ``\textit{Your Son is a \textbf{Retarded} f*ggot like his Cowardly Daddy}'', respectively.
Additionally, Directed hate speech has the highest proportions of being\_obligated frames (0.02, $p < 0.01$) in comparison to generalized hate (0.014) and general tweets (0.013). A tweet that demonstrates this is \textit{``@usr your a f*ggot and should suck my tiny c*ck block me pls''}. 

\vspace*{0.05in}
\noindent
\textbf{Generalized hate speech evokes concepts such as \emph{People by religion, Killing, Color, People,} and \emph{Quantity}}.
Figure~\ref{fig:semafor_barchart} shows that generalized hate has the highest proportion of frames related to People (0.033 vs 0.02 for Directed hate and 0.025 for Gen-1\%, $p < 0.01$), People\_by\_religion (0.06 vs 0.002 for Directed hate and 0.001 for Gen-1\%, $p < 0.01$), Killing (0.03 vs 0.006 for Directed hate and 0.003 for Gen-1\%, $p < 0.01$), Color (0.02 vs 0.012 for Directed hate vs 0.004 for Gen-1\%, $p < 0.01$), and Quantity (0.042 vs 0.025 for Directed hate and 0.026 for Gen-1\%, $p < 0.01$).
Example tweets include: ``\textit{@usr @usr @usr Anything to trash this \textbf{black} President!!}''; ``\textit{Why \textbf{people} think gay marriage is okay is beyond me. Sorry I don't want my future son seeing 2 f*gs walking down the street holding hands}''; and ``\textit{@usr how \textbf{many} f*ckin fags did a even get? Shouldnt be allowed into my wallet whilst under the influence haha}''.

\vspace*{0.05in}
\noindent
\textbf{General tweets (Gen-1\%) primarily evoke concepts related to the \emph{Cardinal Numbers} and \emph{Calendric Units.}}
General tweets have been found to have the highest proportion of cardinal numbers (0.03 vs 0.016 for Directed hate and 0.02 for Generalized hate, $p < 0.01$) and calendric units (0.031 vs 0.01 for Directed hate and 0.013 for Generalized hate, $p < 0.01$). Examples include: ``\textit{I LOVE you usr! xxx \textbf{February 20, 2017} at \textbf{05:45AM} \#AlwaysSuperCute}'' and ``\textit{Women's Basketball trails Fitchburg at the half \textbf{39-32}. Chelsea Johnson leads the Bulldogs with \textbf{12}. Live stats link: https://t.co/uRRZosr7Cl}.'' 

\indent As a final step, we analyze the top words that evoked the top $10$ frames in each type. We summarize these results in Figure \ref{fig:semafor_wc}.
In Directed hate speech, we observe the presence of words like \texttt{do, doing, does, did, get, mentions, says}, which evoke the concept of \textsc{Intentional Acts}.  This suggests that Directed hate speech directly and explicitly calls out the action of or toward the target. We also note the presence of \textsc{hindering} words like \texttt{retard, retarded}, which are explicitly used to attack the target entity.  In contrast, Generalized hate speech is dominated by words that evoke \textsc{Killing} (\texttt{kill, murder, exterminate}), words that categorize \textsc{people by religion} (\texttt{jews, christians, muslims, islam}) and words that refer to a \textsc{Quantity} (\texttt{million, several, many}).  This suggests the broad and general nature of Generalized hate speech, which seeks to associate hate with a general large community or group of people.



\section{Discussion and Conclusion}

\noindent \textbf{Social Implications.} The distinction between Directed and Generalized hate speech has important implications to law, public policy and the society.~\citeauthor{wolfson1997hate} raises the intriguing question of whether one needs to distinguish between emotional harm imposed on private individuals from emotional harm imposed on public political figures or from racist/hateful remarks targeted at a general community and no specific individual in particular~\cite{wolfson1997hate}. One position is that according to the First Amendment, one needs to provide adequate opportunities to express differing opinions and engage in public political debate. However, \cite{wolfson1997hate} also notes that in the case of private individuals, the focus shifts towards emotional health and therefore directed/personal attacks or hate speech aimed at a particular individual must be prohibited. According to this position, hate speech directed at a public political figure or a community or no one in particular might be protected. On the other hand, one might argue that hate speech directed at a community has the potential to mobilize a large number of people by enabling a wider reach and can have devastating consequences to society. However, prohibiting all kinds of offensive/hate speech -- Directed or Generalized opens up a slew of other questions with regards to censorship and the role of the government. In summary, this distinction between Generalized and Directed hate speech has widespread and far-reaching societal implications ranging from the role of the government to the framing of laws and policies.

\noindent \textbf{Hate Speech Detection and Counter Speech.} Current hate speech detection systems primarily focus on distinguishing between hate speech and non-hate speech. However as our analysis reveals, hate speech is far more nuanced. We argue that modeling these nuances is critical for effectively combating online hate speech. Our research points towards a richer view of hate speech that not only focuses on language but on the people generating it. For example, we show that Generalized hate exhibits the presence of the ``Us Vs. Them'' mentality~\cite{cikara2011us} by  emphasizing the usage of third person plural pronouns.
Moreover, our results distinguish the different roles intermediaries could develop to deal with digital hate -- one is educating communities to advance digital citizenship and facilitating counter speech~\cite{citron2011intermediaries}. Our study opens the door to research investigating whether different strategies should be designed to combat Directed and Generalized hate.

\noindent \textbf{Conclusion}. In this work, we shed light on an important aspect of hate speech -- its target. We analyzed two different kinds of hate speech based on the target of hate: \textbf{Directed} and \textbf{Generalized}. By focusing on the target of hate speech, we demonstrated that online hate speech exhibits nuances that are not captured by a monolithic view of hate speech - nuances that have social bearing. Our work revealed key differences in linguistic and psycholinguistic properties of these two types of hate speech, sometimes revealing subtle nuances between directed and generalized hate speech. Additionally, our work highlights present challenges in the hate speech domain. One key challenge is the variety of platforms that incubate hate speech other than Twitter. Other challenges include overcoming sample quality issues and other issues associated with Twitter Streaming API as discussed by ~\cite{tufekci2014big,morstatter2013sample}, and the need to move beyond keyword-based methods that have been shown to miss many instances of hateful speech~\cite{saleem2016web}. Despite these challenges, our approach has enabled us to amass a large dataset, which led us to a number of novel and important understandings about hate speech and its usage. We hope that our findings  enable additional progress within  counter speech research.

\bibliographystyle{aaai}
{\footnotesize
\bibliography{p2p_ref_short}}

\end{document}